\documentclass[lettersize,journal]{IEEEtran}
\usepackage{amsmath,amsfonts}
\usepackage{algorithmic}
\usepackage{algorithm}
\usepackage{array}
\usepackage[caption=false,font=normalsize,labelfont=sf,textfont=sf]{subfig}
\usepackage{textcomp}
\usepackage{stfloats}
\usepackage{url}
\usepackage{verbatim}
\usepackage{graphicx}
\usepackage{cite}
\usepackage[colorlinks, linkcolor=blue]{hyperref}
\usepackage{xcolor}
\begin{document}

\title{Language-Guided Face Animation by Recurrent StyleGAN-based Generator}

\author{
Tiankai Hang, 
Huan Yang,
Bei Liu,
Jianlong Fu,
Xin Geng,
Baining Guo
\thanks{
The work is done when Tiankai is a research intern at Microsoft Research Asia.

Tiankai and Xin are with the School of Computer Science and Engineering, and the Key Laboratory of Computer Network and Information Integration, Ministry of Education, Southeast University, Nanjing 211189, China (email: tkhang@seu.edu.cn; xgeng@seu.edu.cn).

Huan, Bei, Jianlong, and Baining are with Microsoft Research Asia (huan.yang@microsoft.com; bei.liu@microsoft.com; jianf@microsoft.com; bainguo@microsoft.com).
}
}

\markboth{
}%
{Shell \MakeLowercase{\textit{et al.}}: A Sample Article Using IEEEtran.cls for IEEE Journals}

\maketitle

\begin{abstract}
Recent works on language-guided image manipulation have shown great power of language in providing rich semantics, especially for face images. However, the other natural information, motions, in language is less explored. In this paper, we leverage the motion information and study a novel task, language-guided face animation, that aims to animate a static face image with the help of languages. To better utilize both semantics and motions from languages, we propose a simple yet effective framework. Specifically, we propose a recurrent motion generator to extract a series of semantic and motion information from the language and feed it along with visual information to a pre-trained StyleGAN to generate high-quality frames. To optimize the proposed framework, three carefully designed loss functions are proposed including a regularization loss to keep the face identity, a path length regularization loss to ensure motion smoothness, and a contrastive loss to enable video synthesis with various language guidance in one single model. Extensive experiments with both qualitative and quantitative evaluations on diverse domains (\textit{e.g.,} human face, anime face, and dog face) demonstrate the superiority of our model in generating high-quality and realistic videos from one still image with the guidance of language. Code will be available at \href{https://github.com/TiankaiHang/language-guided-animation.git}{https://github.com/TiankaiHang/language-guided-animation.git}
\end{abstract}


\section{Introduction}
\IEEEPARstart{H}{igh-quality} visual content creation (\textit{e.g.}, image and video generation) is a long-standing pursuit of the community. Image generation and manipulation have been studied by many research works and have achieved great success in specific domains and complex scenes~\cite{stylegan,stylegan2,biggan,dhariwal2021diffusionbeatsgan,pmlr-v162-glide}. Recent works make a step forward to leverage languages as semantic guidance in manipulation and propose text-guided image manipulation~\cite{paintbyword,xia2021tedigan,styleclip}. However, the other most important information, motions(\textit{e.g.}, ``smiling'' is dynamic process instead of a static state), in language is neglected. Inspired by this, we study a novel task, language-guided face animation, that leverages motion information from language and animates a still face image. With the help of language, our results could achieve a more vivid illustration of the face and benefit a broad range of real-life scenarios, such as film industry, virtual reality, games, etc.

It should be addressed that it is nontrivial to extract both motion and semantic information from language and animate a still image into a high-quality and smoothed video frame sequence. 
Directly applying existing methods to our task, language-guided face animation, is challenging. StyleCLIP~\cite{styleclip} is a state-of-the-art method that aims for text-driven image manipulation by pre-trained StyleGAN~\cite{stylegan,stylegan2} and CLIP~\cite{CLIP}. A straightforward way to get the motion is interpolating between the source code and edited code from StyleCLIP~\cite{styleclip}. It relies on the last optimized code and may lead to unwanted artifacts,\textit{e.g.}, the move or disappearance of hair. Also, the interpolation process is not learnable and the uniform motion changes make the video unrealistic. Besides, CLIP loss is designed to optimize the similarity between images and a given specific text~\cite{styleclip,paintbyword} which means multiple models should be trained for different input texts. Pumarola \textit{et al.}~\cite{pumarola2018ganimation} propose to learn a motion space and sample target motion from it. MoCoGAN-HD~\cite{mocogan-hd} learns the motion in an implicit way by LSTM encoders. However, the motion space learned is limited and fails to model precise face motions (e.g., face expressions) that are beyond action units(AUs). In this case, the generated videos are not guaranteed to be semantically matched with the input text.

\begin{figure*}[!ht]
    \centering
    \includegraphics[width=\textwidth]{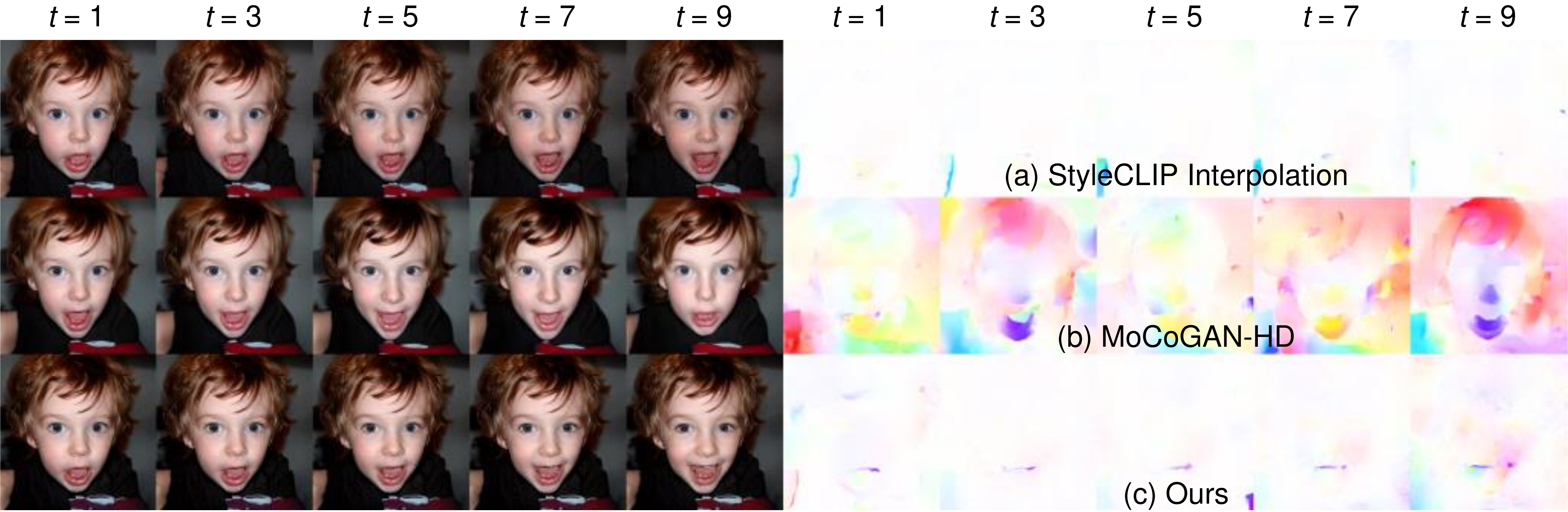}
    \caption{Frames generated from a face image indicating ``smiling'' by three methods: StyleCLIP interpolation,  MoCoGAN-HD, and ours. Optical flow~\cite{teed2020raft} visualizes the motion between adjacent frame. StyleCLIP interpolation keeps the consistency well while the motion is limited. Results from MoCoGAN-HD show large motion with unexpected jitter. Our method not only keeps the consistency but also presents the motion.}
    \label{fig:teaser}
\end{figure*}

The second challenge is how to balance temporal consistency and rich yet meaningful motions, which are both significant to the quality of a video. Direct interpolation~\cite{pumarola2018ganimation} is usually used to keep the temporal consistency while making the synthesized videos less realistic, as shown in Fig.~\ref{fig:teaser}(a). LSTM-like encoders are used for motion generation~\cite{mocogan,mocogan-hd} while they are easy to overfit and sensitive to inputs. Moreover, the jittering between video frames caused by those encoders makes the video lack coherence, as illustrated in Fig.~\ref{fig:teaser}(b). To make sure the result video to have meaningful motions, existing face animation methods rely heavily on action units~\cite{pumarola2018ganimation,imagetovideo-3ddynamics,first-order-image-animation} or motion patterns learned from the training data~\cite{mocogan-hd}. Action units (AUs)~\cite{ekman1976measuring} describe the defined structure for human faces. Recent methods learn a motion space and sample motion from it~\cite{mocogan,mocogan-hd,hongwei-animation}. However, it is difficult to ensure the sampled motion matches the semantics of the language.

To ease the challenges and achieve both per-frame's high quality and temporal consistency, we present a simple yet effective framework for language-guided face animation. Our framework consists of several key components: a cross-modal image-text encoder, a recurrent motion generator, visual and text-aware motion mapper, and a pre-trained image synthesizer (\textit{i.e.}, StyleGAN~\cite{stylegan2}). We take a pair of image and text as input and get the embedding by cross-modal encoders. The text embedding is fed to the \textit{recurrent motion generator} to extract motion and then mapped by a text-aware motion mapper to StyleGAN's $\mathcal{W}+$ space. As illustrated in StyleCLIP~\cite{styleclip}, we first invert the input image to a latent code by a encoder designed for StyleGAN~\cite{tov2021e4e}. The visual embedding is processed by a visual mapper and then combined with the inversion code to get the source content code. The content code and motion code are sent to StyleGAN for per-frame generation. We introduce a \textit{contrastive loss} to facilitate our model's diversity, and a latent path regularization loss to ensure the temporal consistency. 
Facial expressions contains rich semantic information and can be used for emotional analysis~\cite{buitelaar2018mixedemotions-tmm}.
We collect descriptions from website, especially those related to facial expressions, and conduct experiments on the CelebA-HQ dataset. We also explore the generalization and robustness of our framework on AFHQ-Dog~\cite{stylegan2}, AnimeFaces~\cite{animeface}, and MetFaces~\cite{stylegan-ada}. The visual results in Fig.~\ref{fig:teaser}(c) demonstrate the advantage of our method in generating vivid face animations with meaningful motions with the help of languages.

Our contributions are summarized as follows:
\begin{itemize}
    \item We are the first to propose the language-guided face animation task, which aims to synthesize high-quality face animations under semantic and motion control by languages.
    \item We design a simple and effective framework based on a recurrent motion generator and a pre-trained StyleGAN. We introduce a contrastive loss to enable our model to handle different text inputs,
    an identity regularization item to keep the face identity, and a path regularization item to ensure smooth motion transition. 
    \item We conduct extensive experiments on human faces, animes, works of art, and dogs. Considering per-frame's high quality, temporal consistency of the whole video, and the alignment between videos and text, our method achieves superior results than state-of-the-art methods.
\end{itemize}

The rest of the paper is organized as follows: We review previous and related works in Section~\ref{sec:related-work}. The whole framework for language-guided face animation is presented in Section~\ref{sec:method}. Experimental results, analysis, and discussion are shown in Section~\ref{sec:exp}. We finally conclude the work in Section~\ref{sec:conclusion}.

\section{Related Work}\label{sec:related-work}

\noindent
\textbf{Language Guided Visual Manipulation.} 
Visual manipulation has been extensively studied and applied across various domains, including style transfer~\cite{johnson2016perceptual,gatys2016styletransfer,ding-24-style} and image translation~\cite{vid2vid,zhu2017cyclegan}.
General Adversarial Networks(GANs) have made great progress in high-quality face, animal, car generation~\cite{stylegan,stylegan2} and complex scene generation~\cite{biggan}. The pre-trained generators, \textit{e.g.}, StyleGAN~\cite{stylegan,stylegan2}, have good disentangled space. A photorealistic image could be generated by sampling a point from pre-trained latent space. The image could be edited by changing the point's moving directions and step length~\cite{shen2020interfacegan,shen2021closedform}. Strong pre-trained cross-modal models, \textit{e.g.}, CLIP~\cite{CLIP}, have been widely used for text-driven image manipulation. Bau \textit{et al.}~\cite{paintbyword} mask an area of interest, apply non-gradient methods to create semantic paint based on open full-text descriptions. Xia \textit{et al.}~\cite{xia2021tedigan} utilize an inversion module designed for StyleGAN, a visual-linguistic similarity learning module, and optimize for each instance. Another work StyleCLIP~\cite{styleclip} proposes three methods for text-driven image editing: latent vector optimization, latent mapper for specific text, and global directions in StyleSpace. 
Diffusion models have demonstrated strong power on content creation~\cite{pmlr-v162-glide,ramesh2022dalle2,saharia2022imagen}. They mask then create with the text guidance.
With CLIP's~\cite{CLIP} supervision, manipulation could happen in space editing~\cite{shi2021spaceedit}, hair editing~\cite{wei2021hairclip}, domain adaptation~\cite{stylegan-nada}, and style transfer~\cite{kwon2021clipstyler}. In our work, we adopt the pre-trained generator~\cite{stylegan2} to get high-quality frames and a cross-modal model to measure the similarity between the frame and given language.

\noindent
\textbf{Video Synthesis.} 
The task aims to generate a frame sequence with temporal consistency. Videos could be generated in an unconditional way or a way conditioned on given images or videos~\cite{mocogan,mocogan-hd,ho2022videodiffusion,ruan2024univg}. For \textit{stochastic video synthesis}, Tulyakov \textit{et al.}~\cite{mocogan} propose to generate a video from a sequence of vectors with fixed content part and stochastic motion part. MoCoGAN-HD~\cite{mocogan-hd} learns motion with LSTM-like architecture and relies on pre-trained image generator to synthesize high-resolution frames(up to $1024\times 1024$). 
Recently, diffusion models are applied for video synthesis~\cite{ho2022videodiffusion}.
\textit{Image to video synthesis} learns a mapping from the  image to an animated video. Dorkenwald \textit{et al.}~\cite{cinns} present a conditional invertible neural network (cINN) to model static and other video characteristics. Xue \textit{et al.}~\cite{hongwei-animation} obtain high-quality and realistic landscape by learning fine-grained motion embedding. \textit{Video to video synthesis} aims to transform a source video, \textit{e.g.}, a sequence of segmentation maps~\cite{vid2vid}, to a target realistic video. Bhat \textit{et al.}~\cite{bhat2004flow} explore motion in input video's texture particles and generate videos with the users' guidance line. Vid2vid~\cite{vid2vid} designs a specific generator and discriminator, coupled with a spatio-temporal optimization objective, to generate high-resolution and realistic videos.
Ye \textit{et al.}~\cite{ye2022audio-tmm} applies dynamic convolution kernels to synthesize the target video from the talking face video and unpaired audio.
Different from existing tasks, we are the first to propose language-guided face animation. 

\textbf{Vision Language Representation Learning.} Image and text have natural correspondence. Multi-modal vision and language representation learning have attracted great attention~\cite{CLIP,align,oscar,flava,yuan2021florence,zellers2021merlot,jiang2018modeling-tmm,hdvila,pixelbert,vilt-icml21}. There are two main kinds of vision language models: cross-modal and multi-modal~\cite{flava}. Cross-modal models focus on the alignment between different modalities and learn representations in a contrastive way~\cite{CLIP,align}. Radford \textit{et al.}~\cite{CLIP} collect 400 million image-text pairs and show impressive zero-shot classification ability. Jia \textit{et al.}~\cite{align} scale up visual and vision-language representation learning with one billion noisy image-text pairs. The pre-trained models learns a joint space of two modalities. Multi-modal models focus on the fusion of visual and textual information and target for multi-modal tasks,\textit{ e.g.}, VQA (Visual Question Answering)~\cite{vilt-icml21,oscar,li2024exploring}. Some recent works~\cite{yuan2021florence,flava} try to build a foundation model for pure vision tasks and vision-language tasks involving both modalities. For recent text-guided visual generation or manipulation tasks, cross-modal models like CLIP~\cite{CLIP} are used to optimize the semantic similarity between a pair of image and text.

\section{Method}\label{sec:method}
\begin{figure*}[ht]
    \centering
    \includegraphics[width=\textwidth]{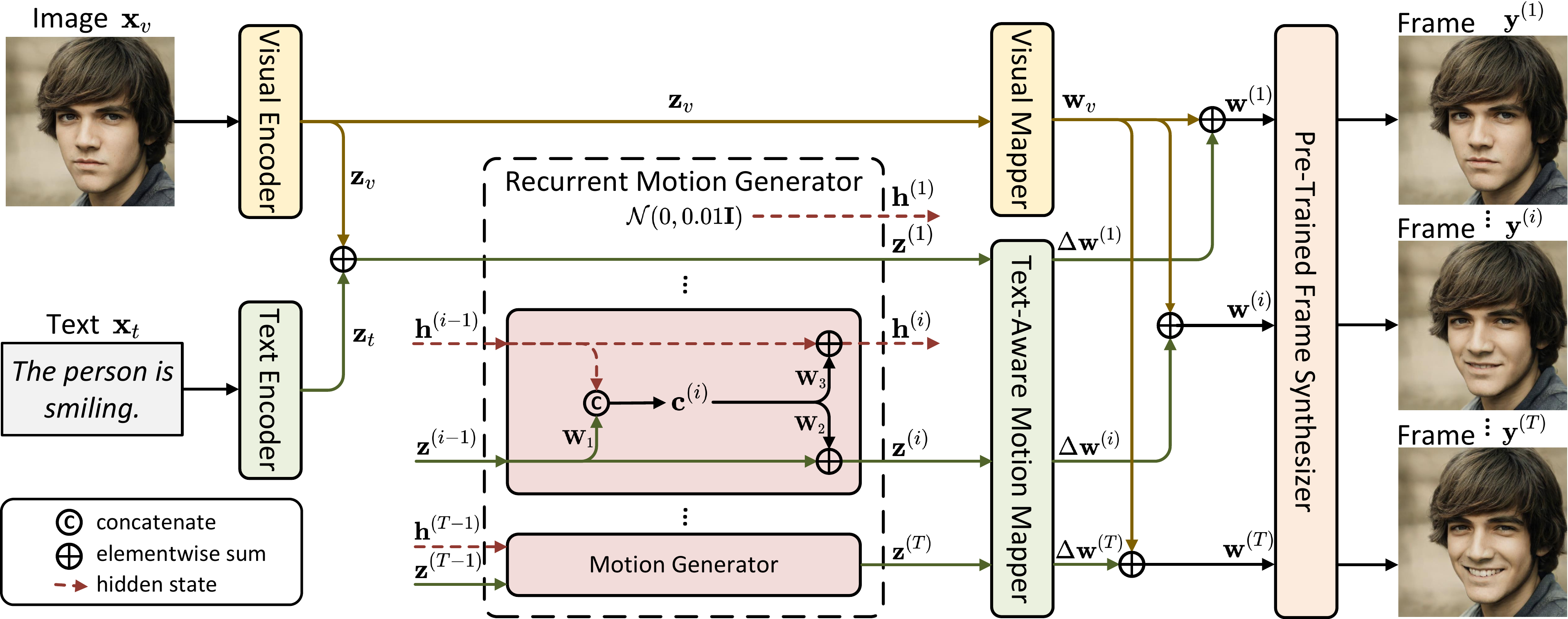}
    \caption{Overview of our proposed framework. We take a pair of image and text as input and embed them by respective encoders. A recurrent motion generator is used to extract motion codes. The visual embedding and motion are processed by visual and text-aware motion mapper. Then the frames are generated by a pre-trained frame synthesizer, \textit{i.e.}, StyleGAN. The synthesizer is fixed during training process}
    \label{fig:framework}
\end{figure*}

We design an effective framework for language-guided face animation. An overview of our approach can be seen in Fig.~\ref{fig:framework}. We first formulate our problem in Section~\ref{sec:formulation}. We then introduce visual-text encoder in Section~\ref{sec:cross-modal-encoder}, the recurrent motion generator in Section~\ref{sec:recurrent}, and the pre-trained high-quality frame synthesizer in Section~\ref{sec:generator}. For diversity and temporal consistency, several loss items are proposed in Section~\ref{sec:lossdesign}.

\subsection{Problem Formulation}\label{sec:formulation}
Our task is to animate the face from a single image with the guidance of natural language. Given a source image $\mathbf{x}_v \in \mathbb{R}^{H\times W \times 3}$ and a text description $\mathbf{x}_t$, our goal is to synthesize a sequence of frames $\{\mathbf{y}^{(i)}\}_{i=1}^{T}$ where $\mathbf{y}^{(i)} \in \mathbb{R}^{H\times W \times 3}$.

\begin{equation}
    (\mathbf{y}^{(1)}, \mathbf{y}^{(2)}, \ldots, \mathbf{y}^{(T)}) = f(\mathbf{x}_v,\mathbf{x}_t, T;\mathbf{\Theta}),
\end{equation}
where the mapping function $f$ consists of several components: 1) visual encoder $E_v$ and text encoder $E_t$, 2) motion generator $M$, 3) visual mapper $H_v$ and text-aware motion mapper $H_t$, 4) a pre-trained generator $G$. All parameters used are $\mathbf{\Theta} = \{\mathbf{\Theta}_E, \mathbf{\Theta}_M, \mathbf{\Theta}_G\}$, where encoders' and mappers' parameters $\{ \mathbf{\Theta}_E, \mathbf{\Theta}_M \}$ are trainable while StyleGAN's parameters $\mathbf{\Theta}_G$ are fixed. 

\subsection{Cross-modal Encoder}\label{sec:cross-modal-encoder}
Benefiting from large-scale vision language pre-training, CLIP~\cite{CLIP} learns a well-aligned space and demonstrates strong power in zero-shot classification. Our cross-modal encoders are directly adopted from OpenAI CLIP~\cite{CLIP}. The visual encoder $E_v$ is a ResNet-50~\cite{resnet}. The text string is tokenized by BPE~\cite{sennrich-etal-2016-bpe} and then encoded by $E_t$ which consists of stacked Transformer layers~\cite{transformer}. The encoded visual and text embedding are denoted as $\mathbf{z}_v$ and $\mathbf{z}_t$.

\begin{equation} \label{equ:encoding}
    \mathbf{z}_v = E_v (\mathbf{x}_v), \qquad \mathbf{z}_t = E_t (\mathbf{x}_t).
\end{equation}

Text descriptions related to expressions contain rich semantic and motion information. 
Then we further feed the text embedding $\mathbf{z}_t$ to our motion generator to get the motion codes.

\subsection{Recurrent Motion Generator}\label{sec:recurrent}
Video synthesis, especially for face animation, can be decomposed into content part and motion part. We design a simple recurrent architecture to obtain the motion codes. To fuse visual-linguistic information and better adapt to different face inputs, motion code is initialized as the sum of visual and text embedding in Equation~\ref{equ:encoding}: $\mathbf{z}^{(1)} = \mathbf{z}_t + \mathbf{z}_v$. The first hidden vector $\mathbf{h}^{(1)}$ is initialized as a random vector sampled from a Gaussian distribution $\mathcal{N}(0, 0.01\mathbf{I})$. The variance for the distribution is empirically set. From the starting state $(\mathbf{z}^{(1)}, \mathbf{h}^{(1)})$, we could generate a sequence of $T$ motion codes in a recurrent way:

\begin{align}
\mathbf{c}^{(i)} &= \sigma (\mathbf{W}_1 \mathbf{z}^{(i-1)} ) \mathbin\Vert  \mathbf{h}^{(i-1)} \label{equ:intermediate}, \\
    \mathbf{z}^{(i)} &= \mathbf{z}^{(i-1)} + r_1 \cdot \sigma (\mathbf{W}_2 \mathbf{c}^{(i)}), \\
     \mathbf{h}^{(i)} &= \mathbf{h}^{(i-1)} + r_2 \cdot \sigma (\mathbf{W}_3 \mathbf{c}^{(i)}),
\end{align}
where $i=2,3,\ldots, T$, $\sigma(\cdot)$ is the activation function, here we use leaky ReLU~\cite{xu2015leakyrelu} with negative slope 0.1. ``$\mathbin\Vert$'' denotes the concatenation operation. $\mathbf{c}^{(i)}$ and $\mathbf{h}^{(i)}$ denote the context vector and hidden state at timestamp $i$. $\{\mathbf{W}^{(i)}\}_{i=1}^3$ are learnable weight matrices.

Existing methods demonstrate that motion residuals help estimate the motion trajectory
 and facilitate the disentangling of motion and content~\cite{zhao2018residual,mocogan-hd}. We get the next code in a residual way to keep relative smooth transition from frame to frame: hidden state $\mathbf{h}^{(i-1)}$ and motion vector $\mathbf{z}^{(i-1)}$ at timestamp $i-1$ are first transformed to a context vector $\mathbf{c}^{(i)}$ and then encoded as motion residuals. During training process, $r_1$ and $r_2$ are set to a constant value 0.2.
During inference, longer sequence generation will introduce the accumulation of artifacts. To mitigate the effects, we set 
$r_1$ and $r_2$ to $\frac{0.2 \cdot (T + 1 - i)}{\#inference~frames}$.
The two values decrease linearly with timestamp $i$.

From motion generator, we can get $T$ latent codes $\{ \mathbf{z}_1, \mathbf{z}_2, \ldots, \mathbf{z}_T\} \subseteq \mathbb{R}^{512}$. They are later added with content code for per-frame generation in a pre-trained synthesizer.

\subsection{Pre-trained Frame Synthesizer}\label{sec:generator}
Training generative models for high-resolution images consume a lot of computing resources and suffers from instability. GANs pre-trained on many domains could achieve amazing results, especially StyleGAN~\cite{stylegan2} on face.
Considering StyleGAN's~\cite{stylegan2} strong ability and the comparison with other methods, we adopt it as our generator. 
To achieve better control, we do the manipulation in StyleGAN's $\mathcal{W}+$ space. 
We first map the visual embedding $\mathbf{z}_v$ in Equation~\ref{equ:encoding} and text-aware motion codes $\{\mathbf{z}^{(i)}\}_{i=1}^T$ to $\mathcal{W}+$ space through the mappers.
\begin{align}
    \mathbf{w}_v &= H_v(\mathbf{z}_v),\\
    \Delta \mathbf{w}^{(i)} &= H_t(\mathbf{z}^{(i)}), i=1,2,\ldots, T.
\end{align}

As illustrated in StyleCLIP~\cite{styleclip}, we add them with the inversion code $\mathbf{w}_{\text{inv}}$ from e4e~\cite{tov2021e4e} to ease the training process. The final latent codes for each frame are represented as:
\begin{align*}
    \mathbf{w}^{(i)} &= \Delta \mathbf{w}^{(i)} + \mathbf{w}_v +  \mathbf{w}_{\text{inv}} \\
    &=: \Delta \mathbf{w}^{(i)} + \mathbf{w}_s, i=1,2,\ldots,T,
\end{align*}
where the source code $\mathbf{w}_s$ represents the content of the face. The pre-trained StyleGAN takes the latent codes $\{ \mathbf{w}^{(i)} \}_{i=1}^T$ as input and generates a sequence of frames $\{ \mathbf{y}^{(i)} \}_{i=1}^T$:
\begin{equation}
    \mathbf{y}^{(i)} = G ( \mathbf{w}^{(i)} ) \in \mathbb{R}^{H\times W\times 3}, i=1,2,\ldots,T.
\end{equation}

\subsection{Loss Design}\label{sec:lossdesign}
Loss design is of great significance to ensure temporal consistency and keep the identity. Following StyleCLIP~\cite{styleclip}, we force final latent codes $\{ \mathbf{w}^{(i)} \}_{i=1}^T$ to be close to the source content code $\mathbf{w}_s$. The regularization item is defined as:

\begin{equation}
    \mathcal{L}_{\text{w-reg}}(\{\mathbf{w}^{(i)}\}_{i=1}^T, \mathbf{w}_s) = \sum_{i=1}^T \lVert \mathbf{w}^{(i)} - \mathbf{w}_s \rVert_2^2
\end{equation}

Karras \textit{et al.}~\cite{stylegan2} introduce path length regularization to encourage that a fixed-step change in $\mathcal{W}$ space leads to a fixed-magnitude change in image. Based on this design, we propose to constrain the pairwise motion change to be similar to obtain a video with temporal consistency.

\begin{equation}\label{equ:path-reg}
\begin{split}
    \mathcal{L}_{\text{path-reg}} (\{\mathbf{w}^{(i)}\}_{i=1}^T) = \sum_{i=1}^{T-2}\lVert \delta^{(i+1)} - \delta^{(i)} \rVert_2^2,\\
    \delta^{(i+1)} = \mathbf{w}^{(i+2)} - \mathbf{w}^{(i+1)}, 
    \delta^{(i)} = \mathbf{w}^{(i+1)} - \mathbf{w}^{(i)}.
\end{split}
\end{equation}

In some CLIP~\cite{CLIP} based manipulation methods~\cite{paintbyword,styleclip,xia2021tedigan,shi2021spaceedit}, the optimization target is to maximize the similarity between one image and the given text. However, optimizing or training a mapping network for one specific text prompt restricts the ability to handle different language inputs. Given a mini-batch of $n$ image-language pairs, where the texts are non-repetitive, we can directly maximize the inner similarity of each pair. We find that such an objective leads to ``average collapse'' where the model tries to balance the effect of different input languages but misses their characteristics. To generalize to different language inputs, we propose to optimize in a contrastive way. We denote $\mathbf{T}$ and $\mathbf{V}$ as the set of visual and linguistic embeddings encoded by CLIP: $\mathbf{T} = \{ \mathbf{t}^{(1)}, \mathbf{t}^{(2)}, \ldots, \mathbf{t}^{(n)} \}, \mathbf{V} = \{ \mathbf{v}^{(1)}, \mathbf{v}^{(2)}, \ldots, \mathbf{v}^{(n)} \}$. The loss is expressed as:

\begin{align}
    \mathcal{L}_{\text{cont}} (\mathbf{V}, \mathbf{T}) &= - \sum_{i=1}^{n} \log \frac{\exp(\mathbf{v}^{(i)}\cdot \mathbf{t}^{(i)} / \tau)}{\sum_{j} \exp(\mathbf{v}^{(i)} \cdot \mathbf{t}^{(j)} / \tau}),
\end{align}
where $\tau$ is the temperature parameter and its default value is 0.07. It is worth noting that $\{ \mathbf{v}^{(i)} \}_{i=1}^n$ are from the last frame of $n$ synthesized video sequences.

Besides, we adopt a LPIPS loss~\cite{johnson2016perceptual} $\mathcal{L}_{\text{lpips}}$ to further improve the perceptual quality. 
All arguments of the loss functions mentioned above are omitted for simplicity and the overall objective is defined as:
\begin{equation}\label{equ:overall-loss}
\begin{split}
    \mathcal{L} &= \lambda_{\text{w-reg}} \mathcal{L}_{\text{w-reg}} + \lambda_{\text{path-reg}} \mathcal{L}_{\text{path-reg}} \\ &+ \lambda_{\text{cont}} \mathcal{L}_{\text{cont}} + \lambda_{\text{lpips}} \mathcal{L}_{\text{lpips}},
\end{split}
\end{equation}
where $\lambda_{\text{w-reg}}$, $\lambda_{\text{path-reg}}$, $\lambda_{\text{cont}}$, and $\lambda_{\text{lpips}}$ are weights to balance losses.

\section{Experiments}\label{sec:exp}
\subsection{Implementation Details}

\subsubsection{Network Architecture} 
We adopt CLIP's~\cite{CLIP} model as our cross-modal encoder, where the visual encoder is ResNet-50~\cite{resnet} and the text encoder is built on Transformer layers. 
The CLIP model is pre-trained on 400M image-text pairs.
The synthesizer $G$ is based on pre-trained StyleGAN~\cite{stylegan2}. The parameters for face generation are converted from officially released checkpoints\footnote{https://github.com/NVlabs/stylegan2.git} and the parameters for dog and anime generation are from MoCoGAN-HD\footnote{https://github.com/snap-research/MoCoGAN-HD.}. The generator pre-trained on high-quality image dataset FFHQ has $\mathcal{W}+$ space with dimension $18\times 512$ and could synthesize images with the resolution of  $1024\times 1024$. The generator pre-trained on AFHQ-Dog or Anime has $\mathcal{W}+$ space with dimension $16\times 512$ and could synthesize images with the resolution of $ 512\times 512 $. 

For face generation, we use real images from CelebA-HQ~\cite{CelebAMask-HQ}.
We first inverse the images through the e4e~\cite{tov2021e4e} network.
For anime faces, art works, and dogs, we do not rely on real images and sample from StyleGAN's latent space as the content part $\mathbf{w}_s$.
Concretely, one point is sampled from StyleGAN's $\mathcal{Z}$ space and then mapped to $\mathbf{w}_s$ by StyleGAN's mapping network.
The visual encoder $E_v$ is removed and the trainable parameters are from text encoder $E_t$, recurrent motion generator $M$, and text-aware motion mapper $H_t$.

The matrices $\mathbf{W}_1,\mathbf{W}_2,\mathbf{W}_3$ used in Section~\ref{sec:recurrent} recurrent motion generator are with the shape $ 384 \times 512  $, $512 \times 768 $, and $384\times 768$, respectively.

\subsubsection{Training Setting}
We rely on the pre-trained generator to synthesize high-quality frames and keep it fixed during training. We optimize the encoders, mappers, and recurrent module with learning rate $ 1 \times 10^{-5} $, $ 1 \times 10^{-3} $, and $ 1 \times 10^{-3} $, respectively. The optimizer used is Adam~\cite{kingma2014adam}, with $ (\beta_1, \beta_2) = (0.0, 0.999) $. 
Loss weights $(\lambda_{\text{w-reg}},\lambda_{\text{path-reg}},\lambda_{\text{cont}},\lambda_{\text{lpips}})$ are set to $(1.0,1.0,0.5,1.0)$.
The learning rates are all kept constant during training. We train on human faces for 2000 iterations. It takes less than an hour to train a model that could handle different texts.

\subsubsection{Datasets} 
For the animation of facial expression, we adopt CelebA-HQ~\cite{CelebAMask-HQ} which contains $30,000$ high-quality face images. They are all aligned and are with the resolution of $1024 \times 1024$. As for dog generation, we do not use any dataset, such as AFHQ~\cite{stylegan2}. All we need is a StyleGAN pre-trained on specific domain. The descriptions of human's facial expressions and dogs' expressions are collected from the website. 

\subsubsection{Baselines}
Our task is new and there are no strictly fair baseline methods for comparison. 
We compare with the most related strong baselines, StyleCLIP~\cite{styleclip} and MoCoGAN-HD~\cite{mocogan-hd}, to demonstrate the effectiveness of our method.

\begin{itemize}
    \item StyleCLIP~\cite{styleclip} is the state-of-the-art text-guided image manipulation method. 
    There are three ways to edit the given image: 
    (a)latent optimization;
    (b)latent mapper;
    (c)global directions.
    \item MoCoGAN-HD~\cite{mocogan-hd} is a recently proposed method for video generation with a pre-trained generator~\cite{stylegan2}. It could be used for directly generating videos from the domain of UCF-101, FaceForensics, and Sky Time-Lapse. It also works well in cross-domain generation, \textit{i.e.}, synthesizing a video with content from the distribution of FFHQ and motion from the distribution of VoxCeleb. Here we adopt the cross-domain generation as our compared baseline.
\end{itemize}

\subsubsection{Metrics}

We evaluate our proposed method on both per-frame quality and video quality. 
Following~\cite{mocogan,mocogan-hd}, we calculate ACD to measure video quality.
Following~\cite{xia2021tedigan}, we calculate FID to measure frame quality.
Besides the two metrics, we also resort to some subjects for human preference.
\begin{itemize}
    \item ACD(Average Content Distance)~\cite{mocogan} is proposed to evaluate the temporal consistency of the video. For face videos, we employ a face recognition network to extract features and calculate the average pairwise $\ell_2$ distance.
    \begin{equation*}
        d_{ACD} = \frac{2\sum_{n=1}^{N} \sum_{i=1}^{T-1} \sum_{j=i+1}^T \Vert \mathbf{f}_{nj} - \mathbf{f}_{ni}  \Vert_2}{N\times T\times (T - 1)} ,
    \end{equation*}
    where $N$ is the number of videos, $T (T > 1)$ is the frame number of each video, $\mathbf{f}_{ni}$ denotes the embedding of $n$-th video's $i$-th frame extracted using an InceptionResNet, which is pre-trained on faces.
    The lower ACD $d_{ACD}$ reflects better temporal consistency.
    \item FID~\cite{parmar2021cleanfid} calculates the distance between distributions of real images and generated images in the embedding space.  The embeddings are from an Inception V3 network pre-trained for classification. A lower FID score reflects higher quality.
    \begin{equation*}
        d_{FID} = \sqrt{\Vert \mu_X - \mu_Y \Vert_2^2 + Tr(\Sigma_X + \Sigma_Y - 2\Sigma_X^{\frac{1}{2}} \Sigma_Y \Sigma_X^{\frac{1}{2}})}
    \end{equation*}
    The images from the source domain $X$ and target domain $Y$ are projected to the embedding space. $\mu_X$ and $\mu_Y$ are the mean vector of embeddings from different domains. $\Sigma_X$ and $\Sigma_Y$ are the covariance matrices.
    The lower FID value $d_{FID}$ reflects the quality of generated images is better.
\end{itemize}

The above two metrics are calculated through the extracted semantic information, which something cannot reflect the quality and fidelity of images. We also conduct \textit{\textbf{user study}} for better evaluation.

We sample 600 images from CelebA-HQ~\cite{CelebAMask-HQ} test set and transform them into animations using our trained model.
We also adopt StyleCLIP's~\cite{styleclip} optimization-based method to get the edited code, and then interpolate in the latent space ($\mathcal{W}+$) to formulate the video. There are 1,200 videos in total (600 from our method and 600 from StyleCILP interpolation).

We ask 24 subjects for user study.
The videos are evaluated in 3 dimensions:
1) Per-frame's quality;
2) Temporal consistency;
3) Video-text alignment.
For each evaluation dimension, one subject needs to choose which one is better.

Each subject is assigned 100 randomly sampled identities.
Each time two videos generated by two methods are presented in a row in random order (A/B testing).
The subjects watch the playing video and make three choices related to the evaluation questions for the better one. 
Each video is evaluated by 4 subjects on average for a more convincing comparison.

For per-frame quality, we count how many 
samples
are rated better than state-of-the-art StyleCLIP. We do the same operation for temporal consistency and video-text alignment. The overall results are presented in Table~\ref{tab:image-quality-comp}. Over three quarters of the participants think our results are better.

\subsection{Comparison with StyleCLIP}\label{sec:styleclip-comp}
A straightforward way to get motion is interpolating between the initial source code and the edited code. 
We compare our method with StyleCLIP's latent optimization method. 
For StyleCLIP, we follow the setting in original paper and set the loss weights $(\lambda_{\text{L2}}, \lambda_{\text{ID}}) = (0.08, 0.05)$ which constrain the $\ell_2$ distance in latent space and keep the face identity. 
Each image is encoded to latent code by the inversion model e4e~\cite{tov2021e4e}, which serves as the initial state. 
Starting from the initial point, it takes 300 steps to optimize the code to increase the semantic similarity between the generated image and the given text. During the process, the StyleGAN and CLIP models are frozen.
We generate 16 frames for comparison.
For StyleCLIP~\cite{styleclip}, we get 16 frames in total by interpolating in latent space.
For our method, we train the model for {2,000} iterations. The model takes the text as input and synthesizes frames we want.

\begin{figure*}[!t]
    \centering
    \includegraphics[width=\linewidth]{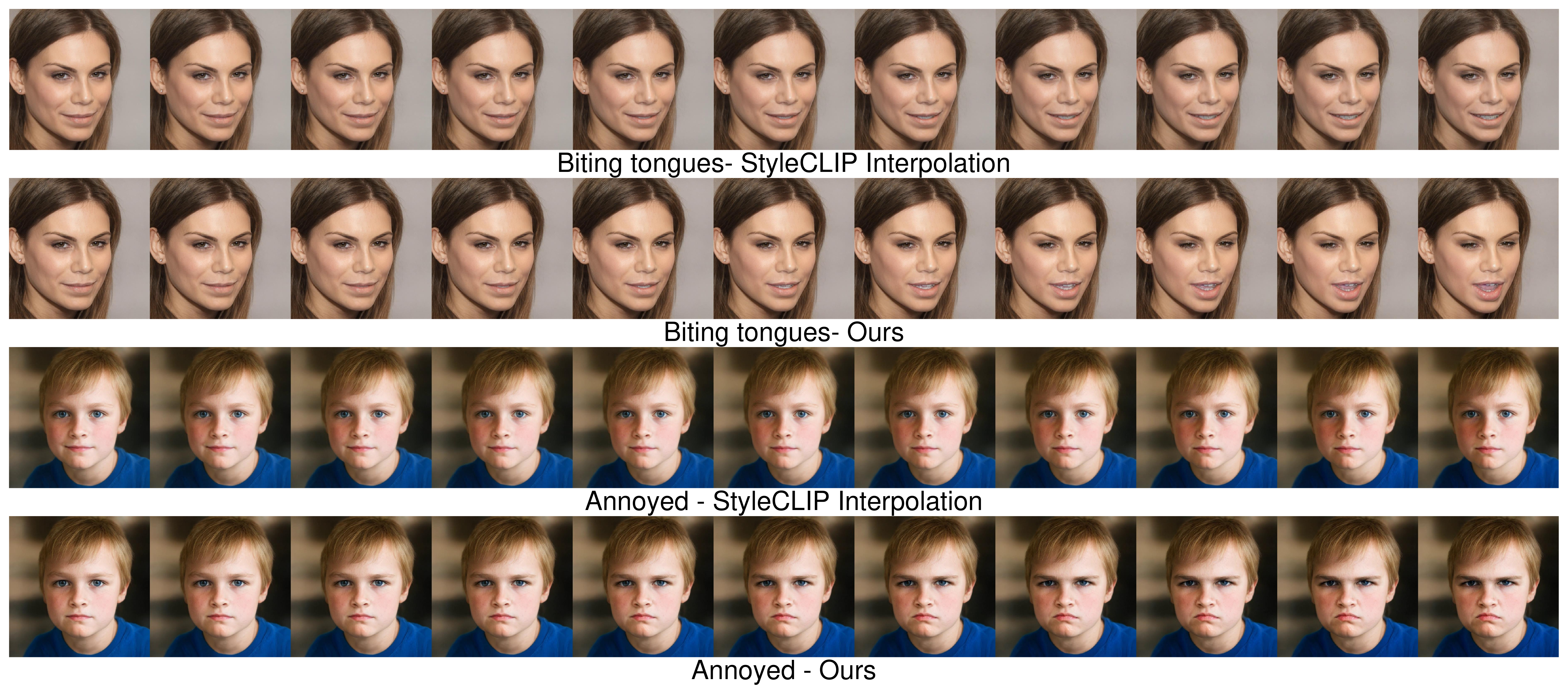}
    \caption{
    Example results of animation using our recurrent motion generator.
    Our results are qualitatively better than StyleCLIP interpolation. The first and third rows are from interpolation, the second and fourth rows are from our model. 
    }
    \label{fig:abl-compare-interp}
\end{figure*}

We randomly sample {600} images from CelebA-HQ's~\cite{CelebAMask-HQ} test set and synthesize videos using our method and StyleCLIP~\cite{styleclip}. The results in Fig.~\ref{fig:abl-compare-interp} show that 1) the edited last frame is superior to StyleCLIP thanks to the learnable recurrent architecture, 2) the transitions in our videos are more natural than StyleCLIP interpolation.
Besides, we calculate the FID score on frames and find 24 subjects to evaluate each video in three dimensions: frame quality(HP-frame), temporal consistency(HP-video), and video-text alignment(HP-alignment). 
Each subject is assigned 100 randomly sampled videos, which means each video is evaluated by 4 subjects on average. 
The FID score and human preference scores are presented in Table~\ref{tab:image-quality-comp}. 
Our model presents more excellence than the baseline.

\setlength{\tabcolsep}{4pt}
\begin{table}[!t]
\begin{center}
\caption{Comparison with StyleCLIP on FID score and human preference. 
$\uparrow$ denotes the higher is the better and $\downarrow$ denotes the lower is the better.
HP-frame, HP-video, and HP-alignment are human preference for frame quality, temporal consistency, and video-text alignment. Almost three quarters of the users prefer our results
}
\label{tab:image-quality-comp}
\begin{tabular}{lcccc}
\hline\noalign{\smallskip}
Method & HP-frame $\uparrow$ & HP-video $\uparrow$ & HP-alignment $\uparrow$ & FID $\downarrow$\\
\noalign{\smallskip}
\hline
\noalign{\smallskip}

StyleCLIP~\cite{styleclip} & {0.272} & 0.237 & 0.238 & {44.22}\\
Ours & {\textbf{0.728}} & \textbf{0.763} & \textbf{0.762} & {\textbf{42.10}}\\

\hline
\end{tabular}
\end{center}
\end{table}
\setlength{\tabcolsep}{1.4pt}

\subsection{Comparison with MoCoGAN-HD}
MoCoGAN-HD~\cite{mocogan-hd} is recently proposed for high-resolution video generation using pre-trained models. The trajectory is randomly sampled and mapped to editing directions in StyleGAN's latent space. The trajectory and editing have no specific semantic meaning, \textit{e.g.}, the trajectory fails to control the person to smile or cry. Their motion is limited to the mouth. We generate {256} videos from the officially released code and pre-trained model for cross-domain FFHQ-VoxCeleb(pre-trained for 12 epochs). 

\begin{figure}[!t]
    \centering
    \includegraphics[width=\linewidth]{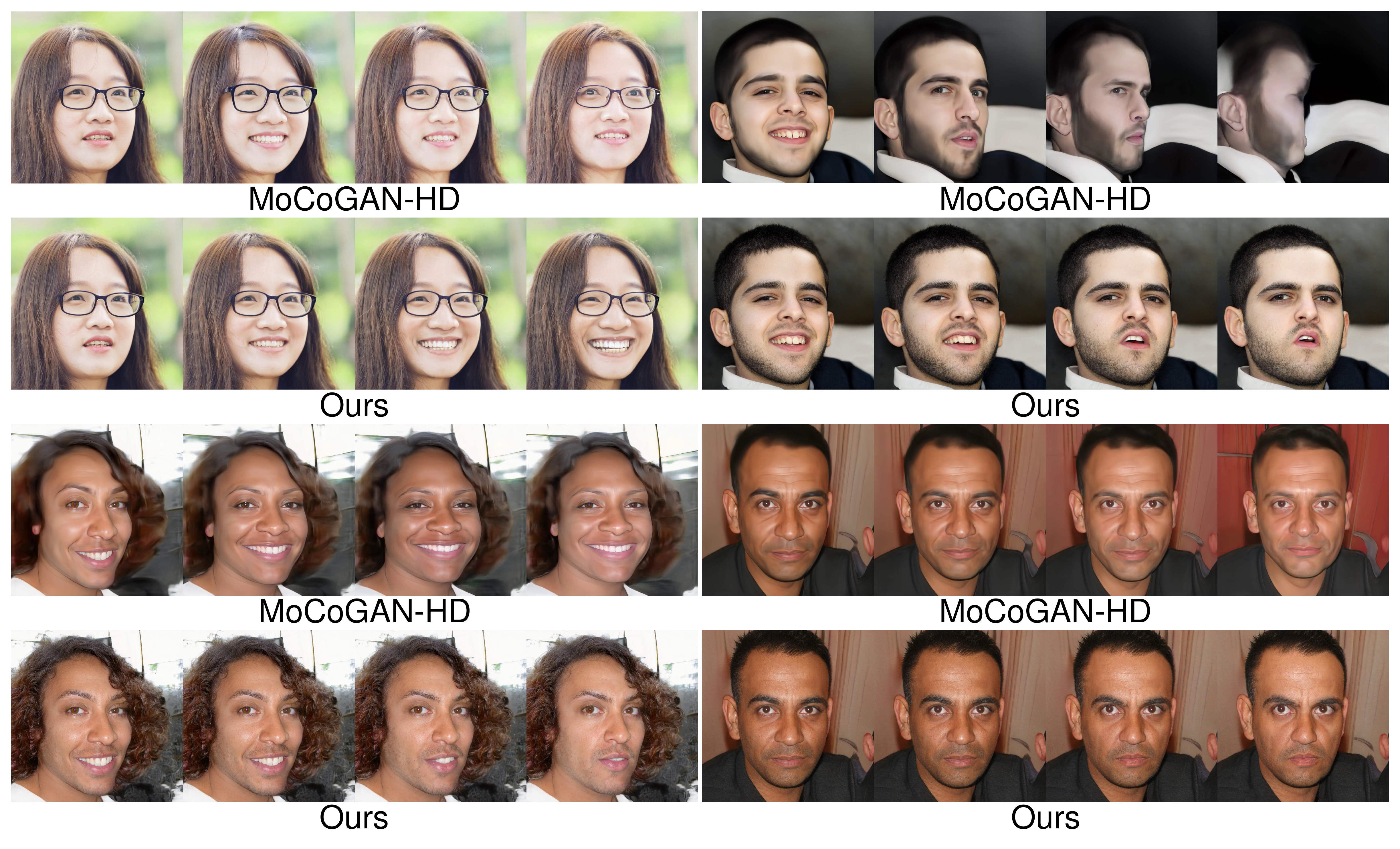}
    \caption{Example results for comparison with MoCoGAN-HD. 
    We sample latent codes from StyleGAN's pre-trained space and synthesize frames using our framework and MoCoGAN-HD.
    Our results show more vivid motion and better temporal consistency than the baseline.
    }
    \label{fig:comp-mocogan}
\end{figure}

In Fig.~\ref{fig:comp-mocogan}, we compare our method with MoCoGAN-HD~\cite{mocogan-hd}. The results of the baseline are with the motion from the randomly sampled trajectory. Trained on VoxCeleb, the motion is limited and focuses on the mouth, which is lack of semantic meaning. The LSTM encoder is unable to keep the identity consistent through all the frames. However, our approach could generate consistent frames with expected effect from the same content code, \textit{e.g.}, ``The person is smiling'' at the left side and ``The person is angry'' at the right side on the second row. In the last row, the results are generated with text ``stunned'' and  ``disgusted''. From the results, we could see that our generate videos are with better temporal consistency and richer semantics than our strong baseline MoCoGAN-HD.

FVD is one metric for video generation evaluation, but it needs ground-truth videos while we only rely on the image dataset.
Following prior works related to video generation~\cite{mocogan,mocogan-hd}, we calculate the \textit{ACD(Average Content Distance)}. Our synthesized videos are with the ACD value of {0.258}, which is {0.193} better than MoCoGAN's 0.451. The results suggest that our model could synthesize videos with better temporal consistency.

\subsection{Ablation Study}
Several key components contribute to the effectiveness of our framework. We have compared with StyleCLIP interpolation method in Section~\ref{sec:styleclip-comp} to demonstrate the strength of our recurrent motion generator.
In this section, we study the effectiveness of our loss design. 

\subsubsection{Loss Design} 
\begin{figure}[!t]
    \centering
    \includegraphics[width=\linewidth]{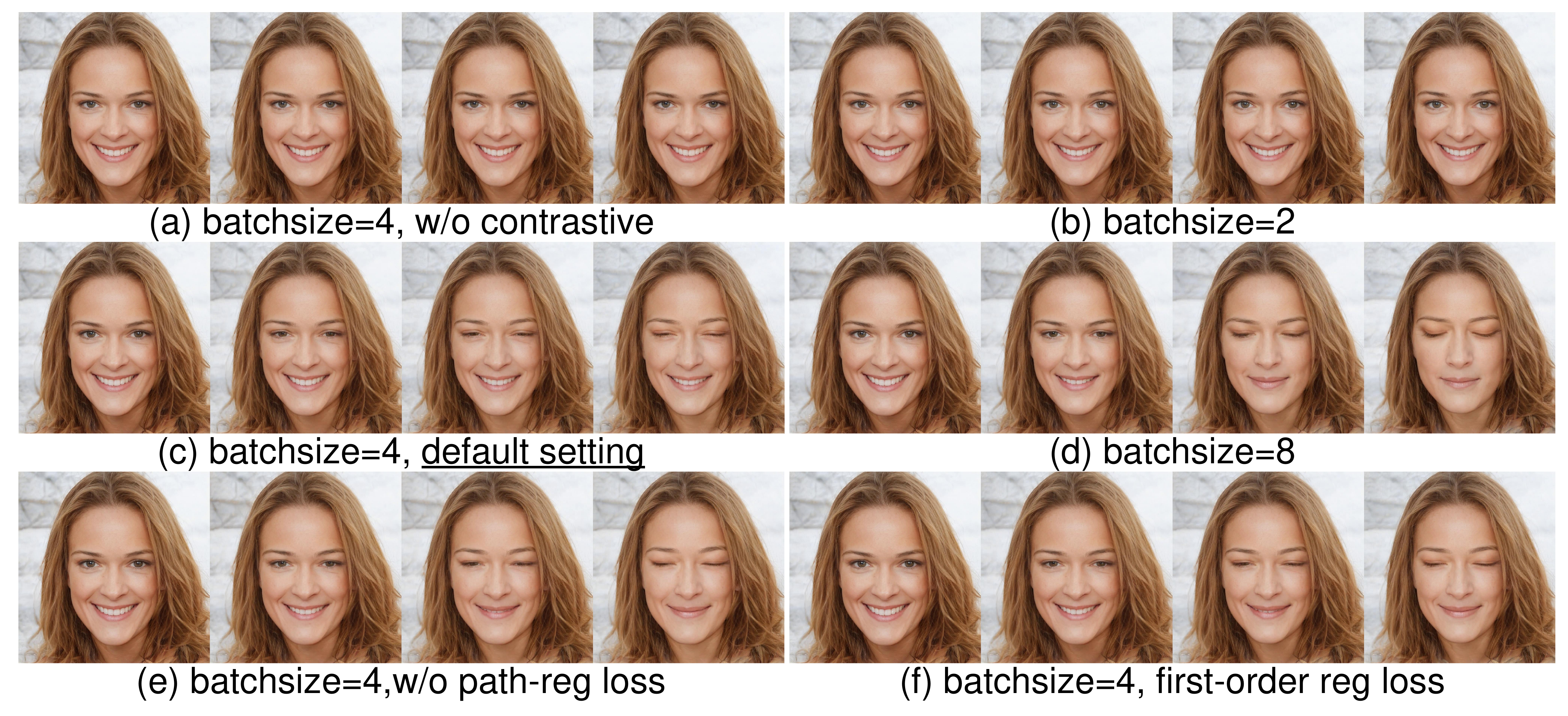}
    \caption{
    Example results for loss ablation study. The guidance is ``closing eyes''.
    (a)-(d) demonstrate the superiority of our contrastive loss. 
    Comparisons between (c) and (e),(f) illustrate better disentanglement of the path regularization loss. (Better zoom in to see the details)
    }
    \label{fig:abl-loss}
\end{figure}

Our loss designs on \textbf{1)} contrastive supervision between the visual and text\textbf{ 2)} latent path regularization contribute to our final results. To evaluate the effectiveness of the contrastive loss, we conduct experiments on several settings: non-contrastive, batch size $=2,4,8$. The results in Fig.~\ref{fig:abl-loss} show that under non-contrastive or small batch size(=2), the model fails to capture the semantics in texts. When batch size is set to 4 or 8, the model is able to handle different texts. We find that batch size 4 is enough for our framework and set to 4 as our default setting. By removing the path regularization item in Equation~\ref{equ:path-reg} or replacing it with first-order difference, the face could close its eyes but our default setting has better motion disentanglement: the person is closing eyes while keeping the mouth open.

\subsection{More Results}

\subsubsection{Generalizing to Animes, Arts, and Dog Animation} 
\begin{figure*}[!t]
    \centering
    \includegraphics[width=\linewidth]{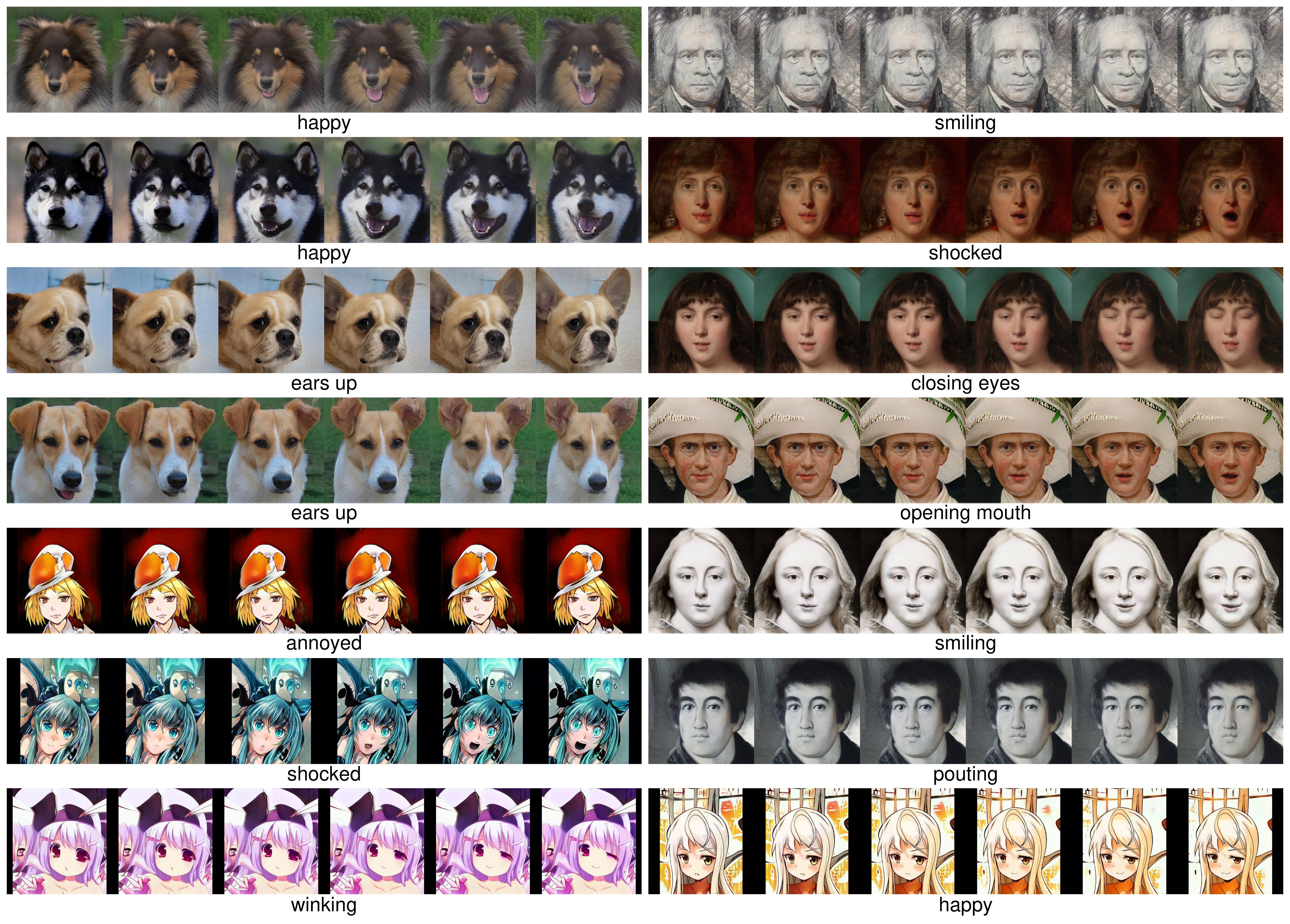}
    \caption{More results for animes, works of art, and dogs.
    Our framework can generalize to different domains of faces.
    }
    \label{fig:general-generation}
\end{figure*}

Our model is not limited to real human face animation but could also generalize to other domains. We do not need real images for training and only rely on the parameters pre-trained on respective datasets. The anime faces and those from works of art share common characteristics with real humans'. The results are shown in Fig.~\ref{fig:general-generation}. We find that MetFaces~\cite{stylegan-ada} share similar characteristics with real human faces, so the texts suitable for CelebA-HQ also work well. Anime faces with simpler structures demonstrate vivid motion changes. There are limited facial expressions for dogs but we still find it works for ``The dog is \{ears up, smiling\}''. The strong results demonstrate the generalization ability of our framework.

\subsubsection{Longer Sequence Generation with Expression Change.} 
\begin{figure*}[!t]
    \centering
    \includegraphics[width=\linewidth]{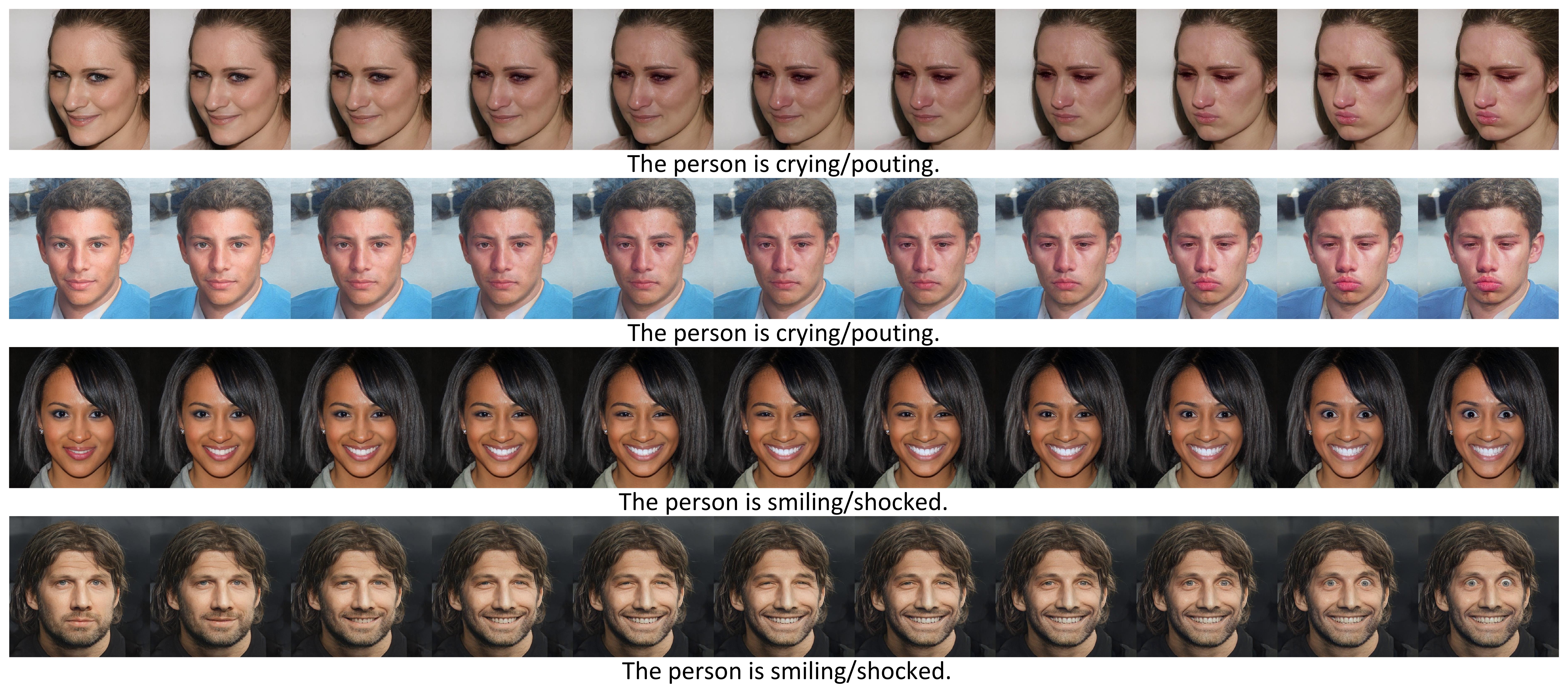}
    \caption{Example results for longer sequence generation(32 frames). The above two cases are with expressions from ``crying'' to ``pouting'', and the below two are from ``smiling'' to ``shocked''. 
    }
    \label{fig:longer-seq}
\end{figure*}

Our model is not only able to synthesize one video with a single expression, but also could also generate longer sequences with expression change. The later expression's initial latent code is from the last frame of the former expression sequence. Results in Fig.~\ref{fig:longer-seq} demonstrate that the identity is well kept and the face gradually changes from one expression to another(\textit{i.e.}, crying to pouting, smiling to shocked) as time goes by.

\subsubsection{Results Beyond Expression}
\begin{figure}[!h]
    \centering
    \includegraphics[width=0.5\textwidth]{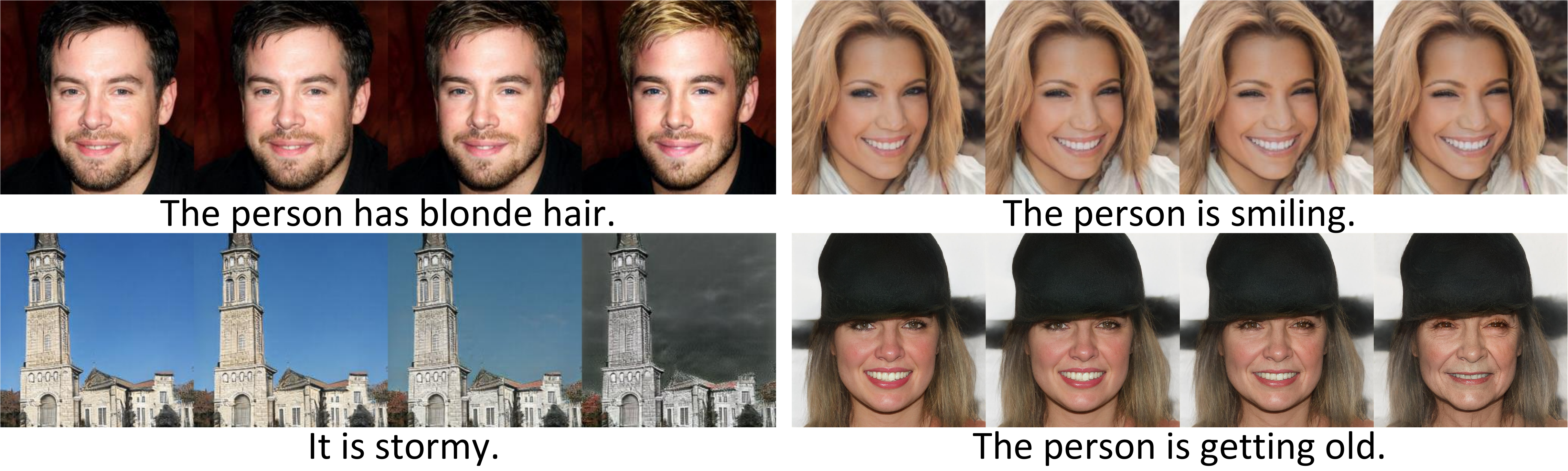}
    \vspace{-0.5cm}
    \caption{More results demonstrate the ability beyond expression. The color change of hair is shown top left; The common expression is shown top right; The weather change is shown bottom left. The age change is modeled as bottom right.}
    \vspace{-0.4cm}
    \label{fig:beyond-expression}
\end{figure}

We conduct experiments on text prompts beyond expression. As shown in Fig.~\ref{fig:beyond-expression}, we can see that 
our method is able to model the color change of the hair, the age change, and the weather change.

\subsection{Discussion}
\begin{figure}[!t]
    \centering
    \includegraphics[width=\linewidth]{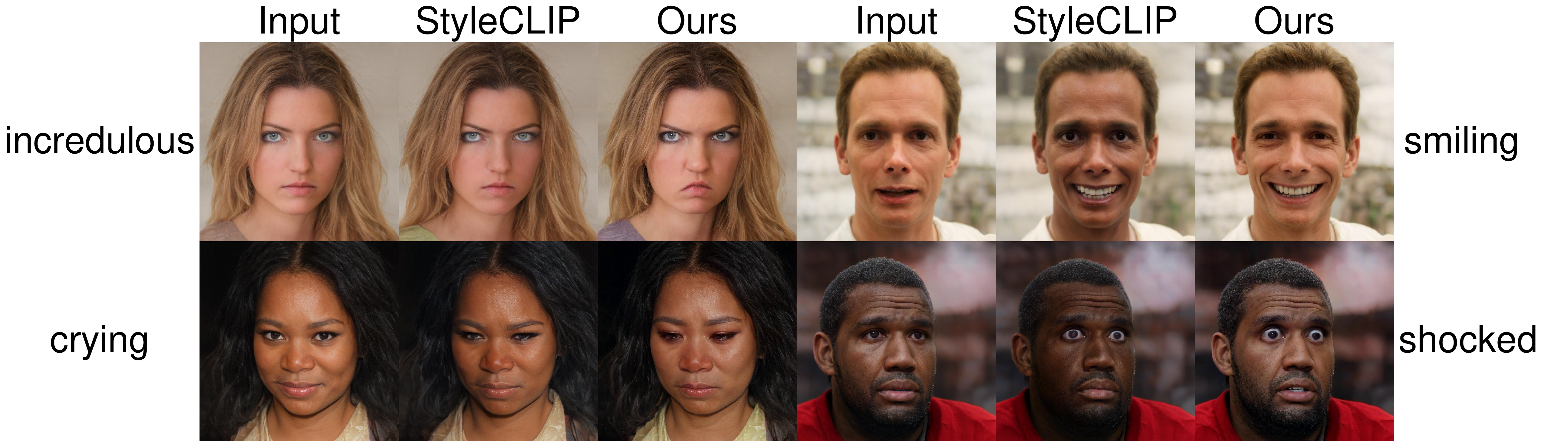}
    \caption{Diverse results of our method. It shows that our method is superior to optimization-based methods in diversity. }
    \label{fig:discussion-diverse}
\end{figure}

\begin{figure}[!t]
    \centering
    \includegraphics[width=0.8\linewidth]{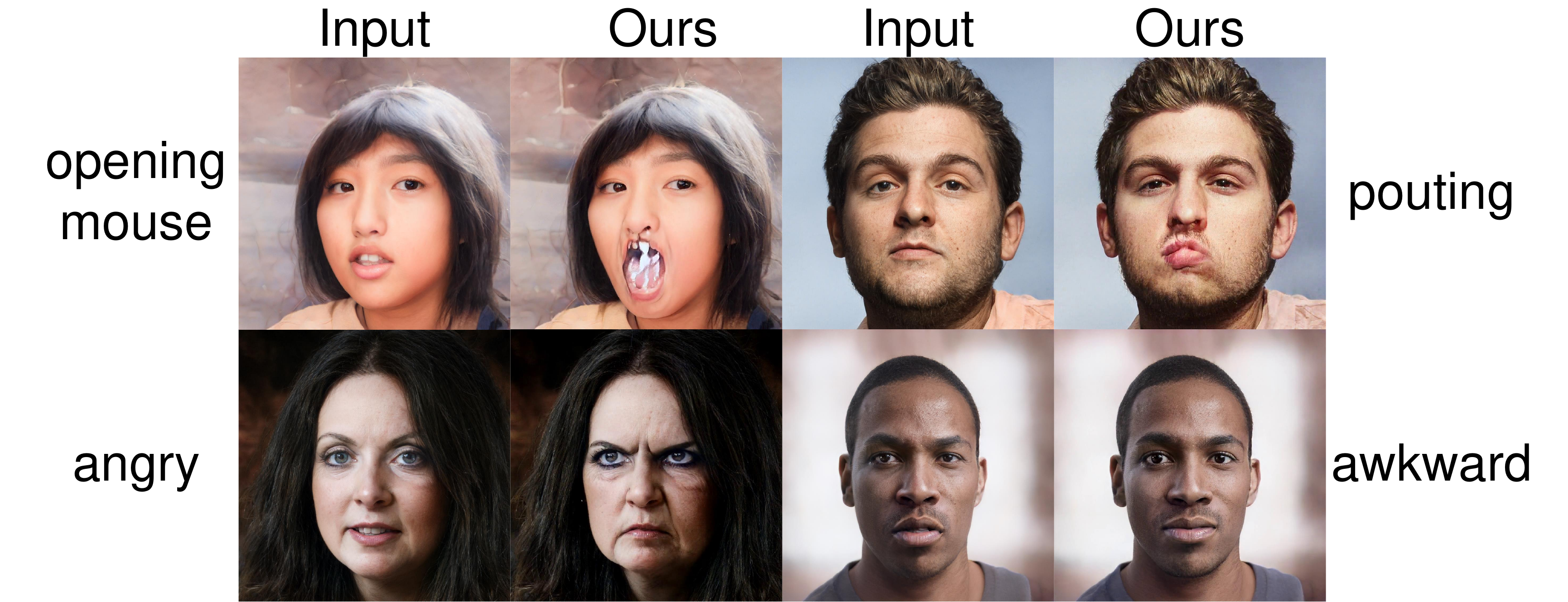}
    \caption{Failure results of our method. It presents some failure cases generated by our model. Expressions with large motion, which have not been seen by StyleGAN, may result in artifacts. Pre-trained CLIP model presents some bias and fails to handle texts with abstract meaning. }
    \label{fig:discussion-failure}
\end{figure}

\subsubsection{Diversity \& Robustness.}
Our framework is able to animate different faces with multiple texts using only one model.
Some results are shown in Fig.~\ref{fig:discussion-diverse}, comparing with the  optimization-based method in StyleCLIP~\cite{styleclip}.
Its global direction method is able to process multiple texts, but we find it collapses (outputs whole black image) to some common seen texts, \textit{e.g.}, ``crying'' and ``stunned'' from the demo website\footnote{https://colab.research.google.com/github/orpatashnik/StyleCLIP/blob/main/\\notebooks/StyleCLIP\_global.ipynb.}.
Our framework is robust to common facial expressions.

\subsubsection{Failure Cases.} Even though our framework works well on many cases, we observe few failure cases, as illustrated in Fig.~\ref{fig:discussion-failure}. CLIP makes it easy to learn some motion patterns, \textit{e.g.}, ``opening mouth'' and ``pouting'', but the generated results are limited by pre-trained StyleGAN. Faces with big open mouths do not follow the distribution of StyleGAN's training data and are difficult to model. 

Pre-trained CLIP introduces bias, which makes ``pouting'' results in age younger and skin whiter, ``angry'' results in age older. Despite large-scale training, CLIP still has difficulties in understanding text with abstract meaning, \textit{e.g.}, ``awkward''.

\subsubsection{The effect of prompt} Prompts are important for language-guided image/video manipulation.
Radford \textit{et al.} also find that prompt engineering plays an important role in zero-shot classification.
The representations of words with or without prompt may be distant in CLIP's~\cite{CLIP} embedding space, which leads to different manipulated results. 
Language with prompts contains more semantic information and reduces uncertainty.
The prompts are important in natural language processing and are worth exploring in our task.

\begin{figure}[!t]
    \centering
    \includegraphics[width=\linewidth]{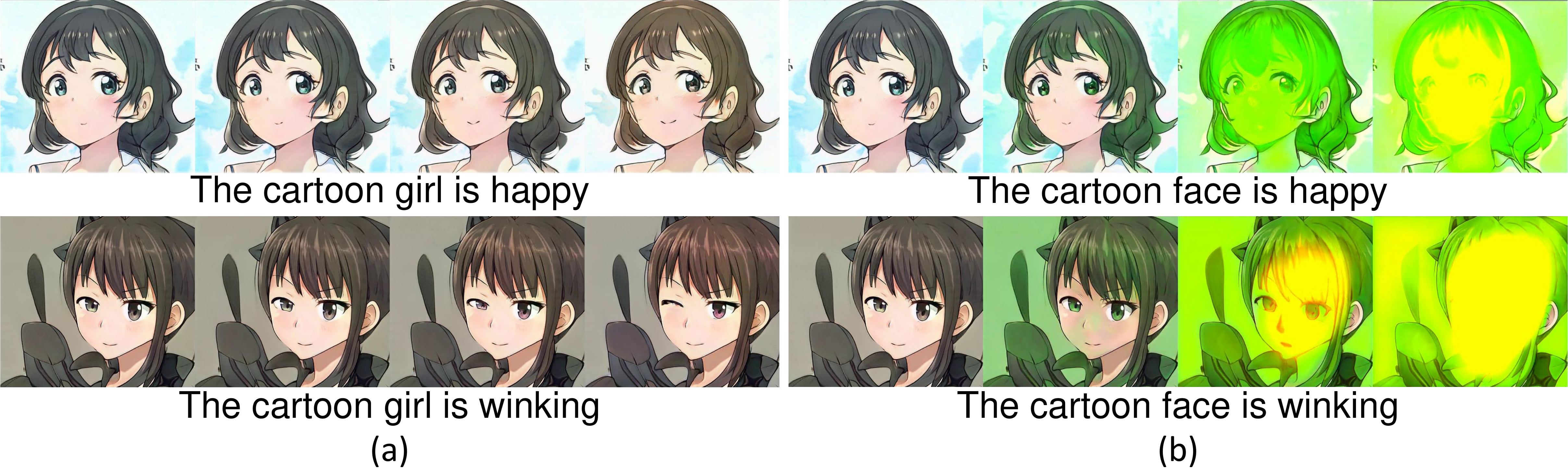}
    \vspace{-0.5cm}
    \caption{Different prompts for anime faces.
    Results in (a) are with the guidance ``The cartoon {\color{blue} girl} is \{happy, winking\}'', while results in (b) are with the guidance ``The cartoon {\color{pink} face} is \{happy, winking\}''. Those with appropriate prompt could animate cartoon faces as expected, while those with different prompts fail to achieve that.}
    \label{fig:pmt-anime}
\end{figure}

\begin{figure}[!t]
    \centering
    \includegraphics[width=\linewidth]{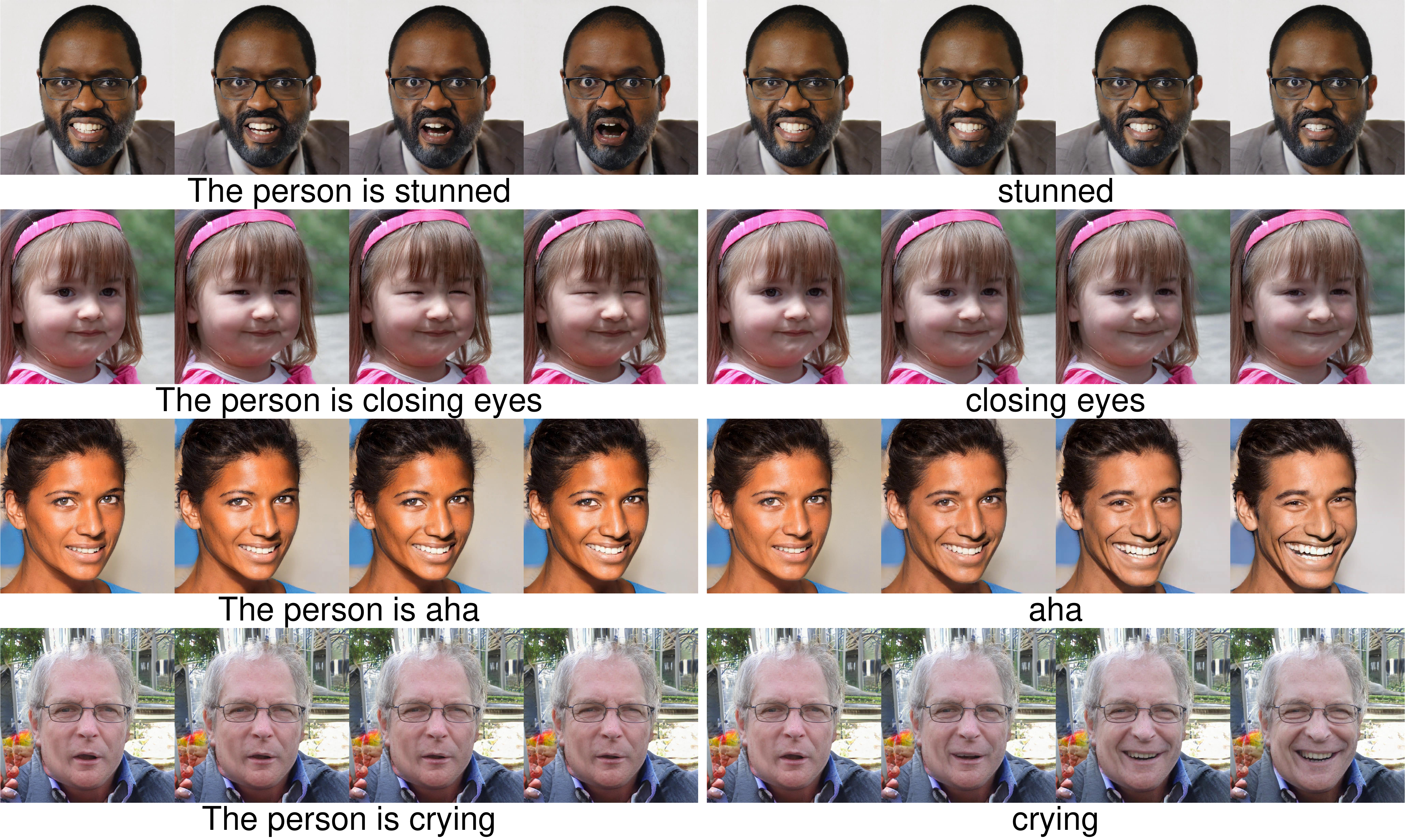}
    \caption{Different prompts for human faces.
    Results in (a) are with the guidance ``The person is \{stunned, closing eyes, aha, crying\}'', while results in (b) are with no prompt. Those with appropriate prompt could help synthesize expected videos, while those without prompts cannot.}
    \label{fig:pmt-face}
\end{figure}

In our experiments, we find that prompt is of great significance to the final results. 
For anime faces, we find the prompt ``\textit{The cartoon girl is}'' and ``\textit{The cartoon face is}'' lead to different results, as shown in Fig.~\ref{fig:pmt-anime}. 
We can see that prompt ``\textit{The cartoon girl is}'' can help synthesize expected results while the prompt ``\textit{The cartoon face is}'' lead to collapse. And for human face animation, the results with prompt ``The person is'' are shown in Fig.~\ref{fig:pmt-face}(a) and results without prompt are in Fig.~\ref{fig:pmt-face}(b). Some cases do not work or present unwanted expressions.

\section{Conclusion}\label{sec:conclusion}
In this paper, we introduce a novel task named language-guided video synthesis and then present a simple yet effective framework for one specific domain - face animation. We propose a recurrent motion generator to extract semantic motion information from language. Together with the powerful pre-trained generator StyleGAN, we are able to synthesize high-resolution videos with language guidance. Three loss functions are designed to ensure better video-language alignment, per-frame's quality, and temporal consistency. We conduct extensive experiments to demonstrate the effectiveness of our method. In future work, we hope to analyze and reduce the biases in pre-trained cross-modality model CLIP for generation. Models pre-trained on image-text pairs are unable to capture the motion and better video-language representations are expected.

\bibliographystyle{IEEEtran}
\bibliography{egbib}

\vfill

\end{document}